\documentclass[runningheads]{llncs}

\usepackage{amsmath,amssymb}       
\usepackage{graphicx}             
\usepackage{changepage}
\usepackage{algorithm,algpseudocode}
\usepackage{tikz}                  
\usetikzlibrary{trees,arrows.meta}
\usepackage{enumitem}
\usepackage{hyperref}

\title{Reasoning and Behavioral Equilibria in LLM-Nash Games: From Mindsets to Actions}
\author{Quanyan Zhu}
\authorrunning{Q. Zhu} 

\institute{Department of Electrical and Computer Engineering, \\ New York University, Brooklyn, NY, USA\\
\email{qz494 @nyu.edu}}

\begin{document}

\maketitle

\vspace{-3mm}\begin{abstract}
We introduce the LLM-Nash framework, a game-theoretic model where agents select reasoning prompts to guide decision-making via Large Language Models (LLMs). Unlike classical games that assume utility-maximizing agents with full rationality, this framework captures bounded rationality by modeling the reasoning process explicitly. Equilibrium is defined over the prompt space, with actions emerging as the behavioral output of LLM inference. This approach enables the study of cognitive constraints, mindset expressiveness, and epistemic learning. Through illustrative examples, we show how reasoning equilibria can diverge from classical Nash outcomes, offering a new foundation for strategic interaction in LLM-enabled systems.
\end{abstract}

% \begin{abstract}
% This paper proposes a game-theoretic framework that models the interaction between prompt engineers and large language models (LLMs) as a two-player extensive-form game and couples it with a Rapidly-exploring Random Trees (RRT) search over prompt space. The attacker incrementally samples, extends and tests prompts, while the LLM chooses to accept, reject or redirect, leading to terminal outcomes of Safe Interaction, Blocked, or Jailbreak. By embedding RRT exploration inside the game, we capture both the discovery phase of jailbreak strategies and the strategic responses of the model. The resulting tree offers a principled foundation for evaluating and hardening LLM guardrails. % 70–150 words 
% \end{abstract}
%\keywords{Prompt–LLM Interaction \and Game Theory \and Jailbreaking \and RRT}

\section{Introduction}

Large Language Models (LLMs) fundamentally reshape decision-making in multi-agent systems, particularly in strategic contexts such as negotiation, cooperation, and adversarial interaction. Traditional game-theoretic formulations typically assume that agents select actions by explicitly solving a utility maximization problem; i.e., choosing the action that maximizes expected utility given the available information and beliefs about other agents' strategies. While this optimization-based approach is grounded in classical decision theory and supports rigorous equilibrium analysis, it often rests on strong assumptions despite efforts to relax them: perfect rationality, complete knowledge of the game structure, and cognitive symmetry across agents. These assumptions, however, are seldom satisfied in real-world settings involving uncertainty, incomplete information, and heterogeneous cognitive capacities.

LLMs offer a new paradigm that relaxes these assumptions. Rather than mapping information to actions via explicit optimization, LLM-based agents generate responses conditioned on observed context and internal latent representations. These representations are high-dimensional and implicitly encode elements such as intent, perception, belief, and anticipation of others' reasoning. Importantly, the reasoning process itself is embedded in the generative capabilities of the LLM, and is influenced by natural language prompts that guide the model’s internal deliberation. In contrast to classical models that treat reasoning as a hidden byproduct of optimization, LLMs make the reasoning process explicit, interpretable, and manipulable.

This shift enables several new directions of inquiry. First, it facilitates the analysis of \emph{bounded rationality} by revealing the intermediate reasoning steps that lead to potentially suboptimal or biased behavior. Second, it invites the design of new solution concepts, such as reasoning-level equilibria, that account for the cognitive scaffolding underlying behavior. Third, it supports a novel class of \emph{epistemic learning processes} in which agents not only adapt their strategies but also refine their reasoning mechanisms over time through prompt engineering or model fine-tuning.

Moreover, LLMs provide a compelling framework for modeling human-like decision-making under uncertainty. Because they are trained on data that reflects diverse forms of human reasoning often under cognitive stress, limited information, or ambiguity, LLMs can naturally replicate heuristic behavior, framing effects, and strategic inconsistencies that are difficult to capture using normative models. As such, LLM-enabled agents serve not only as decision-makers but also as cognitive simulators that bridge the gap between theoretical rationality and observed behavior.

This paper introduces a formal framework termed \emph{LLM-Nash Games}, which extends classical equilibrium concepts to settings where agent policies are generated through prompt-conditioned  LLMs. The framework defines equilibrium not only at the \emph{behavioral level}, specifying which actions are ultimately taken, but also at the \emph{reasoning level}, where those actions are derived via natural language prompts and LLM-based inference. By explicitly modeling both layers, the framework enables the analysis of decision-making in environments characterized by language-based reasoning, epistemic asymmetry, and bounded rationality.

The remainder of the paper is organized as follows. Section~2 formulates the LLM-Nash Game model and relates it to the mental models of agents. Section~3 defines the corresponding equilibrium concept, termed \emph{reasoning equilibrium}, which induces behavioral strategies. Section~4 investigates the relationship between behavioral and reasoning equilibria and studies how the richness of an agent's \emph{mindset} influences decision outcomes. Section~5 presents a case study that illustrates the proposed solution concept in a strategic game. Finally, Section~6 concludes the paper and outlines directions for future work.

\section{LLM-Nash Game Formulation}
 
The \emph{LLM Nash game} extends classical game-theoretic formulations—particularly the standard Nash game by replacing explicit utility maximization with generative reasoning powered by   LLMs. This formulation is especially salient in \emph{agentic AI environments}, where agents must operate under uncertainty, adapt dynamically to feedback, and reason with incomplete or unstructured information. Typical applications include agentic AI negotiation, voting protocols, and security workflows.

Consider a simultaneous-move game involving two players: \( A \) and \( D \). Let \( \mathcal{A} \) and \( \mathcal{D} \) denote the respective action sets of players \( A \) and \( D \). Each player selects a mixed strategy over their action space: player \( A \) selects a distribution \( \mu_A \in \Delta(\mathcal{A}) \) over actions \( a \in \mathcal{A} \), and player \( D \) selects a distribution \( \mu_D \in \Delta(\mathcal{D}) \) over actions \( d \in \mathcal{D} \). We begin with the standard two-person game structure $\mathcal{G} = (\mathcal{A}, \mathcal{D}, u_A, u_D),$ where \( \mathcal{A} \) and \( \mathcal{D} \) are the action spaces of the players, and \( u_A, u_D : \mathcal{A} \times \mathcal{D} \to \mathbb{R} \) are their respective payoff functions. The game is played in a one-shot, simultaneous setting: both players choose their actions independently and without knowledge of the other's choice. The outcome of the game is determined by the joint action profile \( (a, d) \).

Each player also receives an information set, denoted \( I_A \in \mathcal{I}_A \) and \( I_D \in \mathcal{I}_D\), respectively.  The sets \( \mathcal{I}_A \) and \( \mathcal{I}_D \) represent the respective spaces of information that each player may possess during the game.   Information sets contain local observations, past interactions, and contextual knowledge relevant to the strategic decision. In practice, they serve as input to the agents' decision mechanisms. For example, in a cybersecurity context, player \( A \) (the attacker) may observe network traffic and identify vulnerabilities, while player \( D \) (the defender) may monitor system logs and analyze threat indicators.

In the classical formulation, a player's strategy is a mapping from their information set to a mixed strategy. That is, let \( \Gamma_A \) and \( \Gamma_D \) denote the spaces of such mappings (strategy functions), where
\(
\gamma_A : \mathcal{I}_A \to \Delta(\mathcal{A}), \gamma_D : \mathcal{I}_D \to \Delta(\mathcal{D}).
\)
Then, the induced mixed strategies are given by
\(
\mu_A = \gamma_A(\cdot \mid I_A),  \mu_D = \gamma_D(\cdot \mid I_D).
\) 
\begin{definition}[Behavioral Nash Equilibrium]
Consider a two-player game between agents \( A \) and \( D \), where each player's strategy is given by a stochastic policy:
\(
\gamma_A \in \Gamma_A,  \gamma_D \in \Gamma_D,
\)
mapping information inputs to probability distributions over actions. The expected utility for player \( A \), under the strategy profile \( (\gamma_A, \gamma_D) \), is defined as
\(
\bar{u}_A(\gamma_A, \gamma_D) = \mathbb{E}_{a \sim \gamma_A,\, d \sim \gamma_D} \left[ u_A(a, d) \right],
\) 
and similarly for player \( D \),
\(
\bar{u}_D(\gamma_A, \gamma_D) = \mathbb{E}_{a \sim \gamma_A,\, d \sim \gamma_D} \left[ u_D(a, d) \right].
\) 

A strategy profile \( (\gamma_A^*, \gamma_D^*) \) is said to be a \emph{Behavioral Nash Equilibrium} if neither player has an incentive to unilaterally deviate from their strategy. That is, the equilibrium conditions are:
\(
\bar{u}_A(\gamma_A^*, \gamma_D^*) \geq \bar{u}_A(\gamma_A, \gamma_D^*), \ \  \forall \gamma_A \in \Gamma_A,
\)
\(
\bar{u}_D(\gamma_A^*, \gamma_D^*) \geq \bar{u}_D(\gamma_A^*, \gamma_D), \ \  \forall \gamma_D \in \Gamma_D.
\)
\end{definition}
The equilibrium pair \( (\gamma_A^*, \gamma_D^*) \) characterizes the rational behavioral strategies of the agents. At equilibrium, no agent benefits from unilaterally changing their behavior, since each is optimally responding to the other’s strategy.

\subsection*{Definition of LLM Nash Games}
Rather than optimizing over known utility functions, each agent queries an LLM with a structured prompt to generate a probability distribution over available actions. The resulting policies are defined as:
\begin{equation}
\begin{aligned}
\mu_A(a) &= \tilde{\gamma}_A(a \mid I_A, x, \theta) := \mathbb{P}^A_{\texttt{LLM}}[a \mid I_A, x, \theta], \\
\mu_D(d) &= \tilde{\gamma}_D(d \mid I_D, y, \delta) := \mathbb{P}^D_{\texttt{LLM}}[d \mid I_D, y, \delta].
\end{aligned}
\label{eq:llm_policies}
\end{equation}

Here, \( x \in \mathcal{X} \) and \( y \in \mathcal{Y} \) are structured prompts designed by the agents. These prompts encode intent, reasoning heuristics, strategic framing, or a chain of thought to guide the LLM's generative process. They may also serve as \emph{cognitive scaffolding}, shaping how the LLM reasons over the input $I_A$ and $I_D$.  The parameters \( \theta, \delta \in \Xi \) represent the respective \emph{worldviews} of players \( A \) and \( D \), as induced by the underlying LLMs they employ. These worldviews are shaped by the model architecture, training data, and the inductive biases embedded in the LLMs. We refer to the tuples \( \mathfrak{M}_A:=(\mathcal{I}_A, \mathcal{X}, \theta) \) and \( \mathfrak{M}_D:=(\mathcal{I}_D, \mathcal{Y}, \delta) \) as the \emph{mindsets} of the players. Each mindset encapsulates an agent's perception of the environment, their prompt-based reasoning strategy, and their model-specific worldview.

The functions
\(
\tilde{\gamma}^A: \mathcal{I}_A \times \mathcal{X} \times \Xi \rightarrow \Delta(\mathcal{A}), 
\tilde{\gamma}^D: \mathcal{I}_D \times \mathcal{Y} \times \Xi \rightarrow \Delta(\mathcal{D})
\)
denote the generative distributions produced by the LLMs of players \( A \) and \( D \), respectively. These functions are governed by the agents' information inputs, prompts, and the specific LLM models employed, and they yield mixed strategies over actions. Thus, \( \mu_A \) and \( \mu_D \) are the mixed strategies of players \( A \) and \( D \), respectively, as induced by querying their respective LLMs under their current mindsets.

The goal of each player is to construct a reasoning process—encoded as prompts \( x \) and \( y \) for players \( A \) and \( D \), respectively—such that each agent maximizes their expected payoff. In this setting, the strategy mapping traditionally denoted by \( \gamma_A(\cdot \mid I_A) \) is extended to \( \tilde{\gamma}_A(\cdot \mid I_A, x, \theta) \), as defined in Equation~\eqref{eq:llm_policies}. This extended strategy captures not only the informational input \( I_A \), but also the agent’s prompt-based reasoning process \( x \) and the internal model or worldview \( \theta \) induced by the LLM. Note that the mappings \( \tilde{\gamma}_A \) and \( \tilde{\gamma}_D \) are fixed once the underlying foundation LLM model is chosen. That is, the structure of the reasoning-to-action transformation is determined by the pretrained LLM architecture. The goal, therefore, is not to design or learn the mapping \( \tilde{\gamma}_A \) itself, but rather to optimize its operating parameters, such as the input prompts, information, or context embeddings.

The outcome of the reasoning process leads to observable behaviors. These behaviors constitute the observable layer of an agent's decision-making process. In classical models of human behavior, the underlying reasoning process is typically unobservable and must be inferred from external actions.  However, the advent of LLM provides a novel opportunity: the reasoning process itself becomes partially accessible and controllable through prompt engineering and latent representations. This shift enables us to directly probe and manipulate the internal deliberation dynamics of LLM-based agents.

Consequently, the appropriate solution concept should not be defined merely at the behavioral level (i.e., in terms of actions), but rather at the level of reasoning, since it is the reasoning prompt or context that gives rise to behavior. Thus, for an agent (e.g., the attacker), the objective becomes optimizing over the reasoning input \( x \in \mathcal{X} \) that induces favorable behavioral outcomes. Formally, the optimal reasoning is to find within the mindset of the player. 
\[
\max_{x \in \mathcal{X}} \mathbb{E}_{a \sim \tilde{\gamma}_A(\cdot \mid I_A, x, \theta),\ d \sim \tilde{\gamma}_D(\cdot \mid I_D, y, \delta)} \left[ u_A(a, d) \right],
\]
where \( \tilde{\gamma}_A \) and \( \tilde{\gamma}_D \) denote the LLM-based decision mappings (i.e., the foundation models) for the attacker and defender, respectively. These mappings are conditioned on each player's private information \( I_A, I_D \), reasoning prompts \( x, y \), and internal model states \( \theta, \delta \).

Since the foundation LLM is fixed, choosing a specific prompt \( x \) induces a corresponding behavioral policy \( \mu_A^* \), representing the distribution over actions generated by the LLM under optimal reasoning input \( x^* \). Thus, there exists an indirect mapping from the optimal reasoning \( x^* \) to the induced action distribution \( \mu_A^* \). It follows naturally that \( x^* \) should depend on the observed information \( I_A \), as the reasoning prompt must be informed by the agent’s local context.

Accordingly, the reasoning-level decision process can be expressed as a higher-level policy \( \nu^* \colon \mathcal{I}_A \to \mathcal{X} \), which maps the attacker's information set to an optimal reasoning prompt. A similar reasoning policy \( \nu_D^* \colon \mathcal{I}_D \to \mathcal{Y} \) can be defined for the defender.

\begin{remark}
In the above, we have constructed a mental model for an agent enabled by an LLM, mirroring the core structure of human cognition. The cognitive process of such an agent begins with the intake of information—denoted by \( I_A \) and \( I_D \)—which includes environmental signals, private observations, and contextual cues. This intake stage is analogous to \emph{sensory perception} in human cognition, where raw stimuli are collected for further processing. The next phase centers on \emph{reasoning}, where the agent conducts internal mental operations over a form of working memory. This process is shaped by the agent’s past experiences (implicitly encoded in the LLM's pretraining) and is guided by structured prompts and contextual inputs \( x \) and \( y \). These prompts serve to activate specific patterns of reasoning within the LLM, enabling capabilities such as inference, analogy, abstraction, and counterfactual thinking. This stage reflects the human cognitive faculties of deliberation and mental simulation.

Finally, the reasoning process culminates in \emph{observable behavior}, such as selecting an action in a strategic setting or generating a communicative response in a negotiation. This structured progression—from information intake to reasoning to behavior—provides a cognitively interpretable decision-making pipeline that stands in contrast to traditional models, which often map observations directly to actions via rigid utility functions or heuristics. This LLM-based cognitive model is consistent with human-centered frameworks of decision-making as described in~\cite{huang2023cognitive,wickens2013engineering}, where cognition is viewed as a sequential interplay between perception, mental processing, and action.

\end{remark}

\section{LLM Equilibrium}

 \emph{LLM-Nash games} introduce a two-tier equilibrium structure grounded in the cognitive modeling capabilities of foundation models. Each agent, rather than directly selecting a policy over actions, constructs a structured prompt that guides an LLM to generate its action distribution. Thus, the agent’s strategy lies in the prompt space, and the resulting behavior is an emergent outcome of LLM-based reasoning.
 
\begin{definition}[LLM-Nash Equilibrium]
Consider a two-player game between an attacker \( A \) and a defender \( D \), where each player is equipped with a fixed LLM that maps information and structured prompts to action distributions. Let the \emph{mindset} of each player be defined as a tuple consisting of their private information, reasoning prompt, and model-specific internal state: \( (I_A, x, \theta) \in \mathcal{I}_A \times \mathcal{X} \times \Xi \) for the attacker, and \( (I_D, y, \delta) \in \mathcal{I}_D \times \mathcal{Y} \times \Xi \) for the defender.

The LLM-induced mixed strategies are given by: $\mu_A(a) = \tilde{\gamma}_A(a \mid I_A, x, \theta), \
\mu_D(d) = \tilde{\gamma}_D(d \mid I_D, y, \delta),$
where \( \tilde{\gamma}_A \) and \( \tilde{\gamma}_D \) are fixed generative policies defined by the agents' respective LLMs. Given this structure, an \emph{LLM-Nash reasoning equilibrium} is a pair of reasoning prompts \( (x^*, y^*) \in \mathcal{X} \times \mathcal{Y} \) such that neither player has an incentive to unilaterally deviate in prompt design. That is, the expected utility of each player, evaluated over the actions generated by their respective LLMs under the equilibrium prompts, is at least as high as under any alternative prompt, holding the other player's prompt fixed. Formally:
\begin{align}
\mathbb{E}_{\substack{
a \sim \tilde{\gamma}_A(\cdot \mid I_A, x^*, \theta) \\
d \sim \tilde{\gamma}_D(\cdot \mid I_D, y^*, \delta)
}} 
\left[ u_A(a, d) \right]
&\geq 
\mathbb{E}_{\substack{
a \sim \tilde{\gamma}_A(\cdot \mid I_A, x', \theta) \\
d \sim \tilde{\gamma}_D(\cdot \mid I_D, y^*, \delta)
}} 
\left[ u_A(a, d) \right],
\quad \forall x' \in \mathcal{X}, \\
\mathbb{E}_{\substack{
a \sim \tilde{\gamma}_A(\cdot \mid I_A, x^*, \theta) \\
d \sim \tilde{\gamma}_D(\cdot \mid I_D, y^*, \delta)
}} 
\left[ u_D(a, d) \right]
&\geq 
\mathbb{E}_{\substack{
a \sim \tilde{\gamma}_A(\cdot \mid I_A, x^*, \theta) \\
d \sim \tilde{\gamma}_D(\cdot \mid I_D, y', \delta)
}} 
\left[ u_D(a, d) \right],
\quad \forall y' \in \mathcal{Y}.
\end{align}

We call the pair $(\mu_A^*, \mu_D^*) \in \Delta(\mathcal{A}) \times \Delta(\mathcal{D})$ induced by $(x^*, y^*)$ is called the LLM-Nash behavioral equilibrium. 

\end{definition}

This equilibrium concept captures the strategic behavior of agents at the level of reasoning, which in turn induces their equilibrium behavior at the action level. In this framework, utility maximization is embedded within the reasoning process itself, encapsulated by the prompts $x$ and 
$y$. Accordingly, each agent seeks to design (or select) a reasoning strategy from which it has no incentive to deviate at equilibrium.

Importantly, rationality in this setting is relocated from the action level to the reasoning level: agents are rational in the sense that they choose the best reasoning process available within their mindset. However, the bounded rationality of agents, manifested through the limitations and inductive biases of LLM-based reasoning \cite{liu2024exploring}, implies that the resulting behaviors may deviate from those predicted by classical utility-maximizing models. That is, although reasoning is optimal given the agent’s mindset, it may still appear suboptimal when compared to the Behavioral Nash Equilibrium defined in Definition~1.

The existence of a reasoning equilibrium requires further examination. When the prompt spaces \( \mathcal{X} \) and \( \mathcal{Y} \) are non-convex or discrete, the existence of a pure-strategy reasoning equilibrium may not be guaranteed. In such cases, the existence must be studied in an extended mixed-strategy space, formally described as follows.

\begin{theorem}[Existence of Mixed-Strategy LLM-Nash Reasoning Equilibrium]
Let \( \mathcal{X} \) and \( \mathcal{Y} \) be finite sets of structured reasoning prompts available to the players $A, D$, respectively. Consider the expected utility functions defined in Definition 2. Let 
\[
U_A(x, y) := \mathbb{E}_{a \sim \tilde{\gamma}_A(\cdot \mid I_A, x, \theta),\ d \sim \tilde{\gamma}_D(\cdot \mid I_D, y, \delta)}[u_A(a, d)],
\]
\[
U_D(x, y) := \mathbb{E}_{a \sim \tilde{\gamma}_A(\cdot \mid I_A, x, \theta),\ d \sim \tilde{\gamma}_D(\cdot \mid I_D, y, \delta)}[u_D(a, d)],
\]
where \( \tilde{\gamma}_A \) and \( \tilde{\gamma}_D \) are fixed LLM-induced policies.

Suppose each player can select a mixed strategy over prompts:
$\sigma_A \in \Delta(\mathcal{X}), \sigma_D \in \Delta(\mathcal{Y}),$
where \( \Delta(\mathcal{X}) \) and \( \Delta(\mathcal{Y}) \) denote the spaces of probability distributions over \( \mathcal{X} \) and \( \mathcal{Y} \), respectively. Then, there exists at least one mixed-strategy reasoning equilibrium \( (\sigma_A^*, \sigma_D^*) \in \Delta(\mathcal{X}) \times \Delta(\mathcal{Y}) \) such that:
\[
\mathbb{E}_{x \sim \sigma_A^*, y \sim \sigma_D^*} [U_A(x, y)] \geq \mathbb{E}_{x \sim \sigma_A, y \sim \sigma_D^*} [U_A(x, y)] \quad \forall \sigma_A \in \Delta(\mathcal{X}),
\]
\[
\mathbb{E}_{x \sim \sigma_A^*, y \sim \sigma_D^*} [U_D(x, y)] \geq \mathbb{E}_{x \sim \sigma_A^*, y \sim \sigma_D} [U_D(x, y)] \quad \forall \sigma_D \in \Delta(\mathcal{Y}).
\]
\end{theorem}

\section{Behavioral  and Reasoning Equilibrium }
Suppose the reasoning strategy of one agent—say, the defender—is fixed, i.e., \( y \) is fixed, and the induced behavior is fixed at \( \mu_D \). Then, it is clear that the rational behavioral equilibrium, as defined in Definition~1, always weakly outperforms the reasoning-level equilibrium for the attacker. More formally, define
\begin{equation}
u^*_A := \max_{\gamma_A \in \Gamma_A} \mathbb{E}_{a \sim \gamma_A,\, d \sim \mu_D} \left[ u_A(a, d) \right] 
= \max_{\mu_A \in \Delta(\mathcal{A})} \mathbb{E}_{a \sim \mu_A,\, d \sim \mu_D} \left[ u_A(a, d) \mid I_A \right],
\label{eq:behavioral_utility}
\end{equation}
\begin{equation}
\tilde{u}^*_A := \max_{x \in \mathcal{X}} \mathbb{E}_{a \sim \tilde{\gamma}_A(\cdot \mid I_A, x, \theta),\, d \sim \mu_D} \left[ u_A(a, d) \right],
\label{eq:reasoning_utility}
\end{equation}

This inequality \( u^*_A \geq \tilde{u}^*_A \) holds because the behavioral decision-making in~\eqref{eq:behavioral_utility} considers a strictly larger feasible set: any policy \( \mu_A \in \Delta(\mathcal{A}) \) is admissible in the behavioral model, whereas in the reasoning-level model~\eqref{eq:reasoning_utility}, \( \mu_A \) must be \emph{induced} by querying the LLM with a structured prompt \( x \in \mathcal{X} \), i.e., \( \mu_A = \tilde{\gamma}_A(\cdot \mid I_A, x, \theta) \). In general, not all distributions in \( \Delta(\mathcal{A}) \) are realizable via LLM reasoning, as the set of LLM-induced policies is constrained by the model architecture and prompt space.

This observation resonates with the classical insight of Blackwell~\cite{blackwell1953equivalent}, who showed that information structures (or experiments) can be ordered by their informativeness. In this context, the LLM reasoning decisions (\ref{eq:reasoning_utility}) can be viewed as a \emph{garbled} or coarsened version of the behavioral decisions (\ref{eq:behavioral_utility}). The set of achievable policies under prompt-based reasoning is effectively a subset of those under fully flexible (behavioral) control, implying a loss in decision quality when reasoning capacity is bounded or structured by the LLM's internal representation.

\begin{theorem}[Utility Gap Under Behavioral vs. Reasoning]\label{thm:utility_gap}
Let \( \mu_{-i} \in \Delta(\mathcal{A}_{-i}) \) be a fixed strategy of the opposing agent \( -i \in \{A, D\} \). Let \( i \in \{A, D\} \) be the agent of interest, with private information \( I_i \), internal state \( \theta \), and access to an LLM that induces a policy mapping \( \tilde{\gamma}_i(\cdot \mid I_i, x, \theta) \) via a structured prompt \( x \in \mathcal{X} \). Then, the agent's optimal expected utility under behavioral-level decision-making (\ref{eq:behavioral_utility}) is at least as large as that under reasoning-level decision-making (\ref{eq:reasoning_utility}), i.e., $u_i^* \geq \tilde{u}_i^*$.
\end{theorem}

Thus, the behavioral decision problem~\eqref{eq:behavioral_utility} can be interpreted as reasoning under an \emph{open mindset}, where any behavior in \( \Delta(\mathcal{A}) \) is achievable . In contrast, the reasoning-level equilibrium operates under a \emph{closed} or constrained mindset \( \mathfrak{M}_A \), determined by the generative capacity of the fixed LLM foundation model and the structure of its prompt. For a given policy \( \mu_A \) or \( \mu_D \), if there exists no reasoning process \( x \in \mathcal{X} \) or \( y \in \mathcal{Y} \) such that \( \mu_A = \tilde{\gamma}_A(\cdot \mid I_A, x, \theta) \) or \( \mu_D = \tilde{\gamma}_D(\cdot \mid I_D, y, \delta) \), respectively, under the corresponding mindsets \( \mathfrak{M}_A \) and \( \mathfrak{M}_D \), then we say that such behavior is \emph{unsupported} by the mindset of the player.

If we regard the optimal reasoning under the open mindset as representing fully rational behavior, then reasoning under a closed mindset corresponds to bounded rationality. The degree of boundedness can be quantified through an ordering over mindsets. Specifically, we say that a mindset \( \mathfrak{M}_A \) is more constrained than \( \mathfrak{M}_A' \) if
\(
\mathfrak{M}_A \subseteq \mathfrak{M}_A'.
\)
This means that the reasoning and policy space accessible under \( \mathfrak{M}_A \) is a strict subset of that under \( \mathfrak{M}_A' \), implying more limited cognitive flexibility or representational power. Beyond this set inclusion, we can also define an ordering based on \emph{strategic expressiveness}, as formalized below.

\begin{definition}[Ordering of Mindsets by Strategic Expressiveness]
Let \( \mathfrak{M}_A = (\mathcal{I}_A, \mathcal{X}, \theta) \) and \( \mathfrak{M}_A' = (\mathcal{I}_A, \mathcal{X}', \theta') \) be two mindsets for agent \( A \),  for a give $I_A\in \mathcal{I}_A$, each inducing a corresponding set of feasible behavioral strategies via the generative map \( \tilde{\gamma}_A \), i.e.,
\(
\mathcal{M}_A := \left\{ \mu_A(\cdot) = \tilde{\gamma}_A(\cdot \mid I_A, x, \theta) \, : \, x \in \mathcal{X} \right\},
\) \(
\mathcal{M}_A' := \left\{ \mu_A'(\cdot) = \tilde{\gamma}_A(\cdot \mid I_A, x', \theta') \, : \, x' \in \mathcal{X}' \right\}.
\)

We say that mindset \( \mathfrak{M}_A \) is \emph{more expressive} than \( \mathfrak{M}_A' \) if
\(
\mathcal{M}_A' \subseteq \mathcal{M}_A.
\)
In this case, agent \( A \) operating under \( \mathfrak{M}_A \) can induce a (weakly) larger set of feasible strategies than under \( \mathfrak{M}_A' \).
\end{definition}

Similarly, we can define an ordering of mindsets for player \( D \). As shown in Theorem~1, greater strategic expressiveness leads to better outcomes in single-agent decision problems. However, in the context of two-player games or strategic interactions, it is not necessarily the case that increased expressiveness guarantees a better outcome for a player. 

This is because the outcome depends not only on an agent's reasoning capacity but also on the strategic responses of the other player. In fact, higher expressiveness may sometimes be exploited by opponents or lead to overfitting to adversarial behavior. Similar phenomena have been observed in the literature on differential games \cite{bacsar2011prices}, games with incomplete information \cite{akerlof1978market,li2023price}, and learning in games \cite{li2022role}, where richer strategy sets do not always yield dominant outcomes.

\section{LLM-Nash Rock-Paper-Scissors: A Case Study}

Consider again the zero-sum Rock-Paper-Scissors (RPS) game between two agents, \( A \) and \( D \), with action sets \( \mathcal{A} = \mathcal{D} = \{\texttt{Rock}, \texttt{Paper}, \texttt{Scissors}\} \). The payoff function \( u_A(a,d) \) returns \( +1 \) if player \( A \) wins, \( -1 \) if they lose, and \( 0 \) for a tie. Player \( D \) receives payoff \( u_D = -u_A \). Suppose that the agents are again constrained to reasoning spaces defined by natural language prompts, with \( \mathcal{X} = \{x_1, x_2\} \) and \( \mathcal{Y} = \{y_1, y_2\} \). Their beliefs about each other's empirical play frequencies are now slightly altered, leading to different induced distributions.

Let player \( A \)'s information be the empirical frequency of D's plays \( I_A = (0.2, 0.3, 0.5) \), indicating that \( D \) tends to play Scissors more often. Let \( D \)'s belief be \( I_D = (0.6, 0.2, 0.2) \), indicating a bias by \( A \) toward Rock. The reasoning prompts are again:
\vspace{-0mm}\begin{align*}
x_1 &: \texttt{``Exploit opponent's bias. Respond to their most frequent move.''} \\
x_2 &: \texttt{``Assume uniform play. Choose uniformly.''} \\
y_1 &: \texttt{``Randomize to avoid predictability.''} \\
y_2 &: \texttt{``Exploit patterns by countering the last move.''}
\end{align*}\vspace{-0mm}
Given the above, the LLM-induced strategies using a chosen LLM become:
$ 
\mu_A(x_1) = (0.2, 0.6, 0.2),  \ \mu_D(y_1) = \left(\tfrac{1}{3}, \tfrac{1}{3}, \tfrac{1}{3}\right), \ 
\mu_A(x_2) = \left(\tfrac{1}{3}, \tfrac{1}{3}, \tfrac{1}{3}\right), \  \mu_D(y_2) = (0.3, 0.4, 0.3).
$ We compute the expected payoffs for player \( A \) under each prompt pair. Using the standard RPS matrix, we find:
\(
\mathbb{E}[u_A(x_1, y_2)] \approx 0.075, \  
\mathbb{E}[u_A(x_2, y_2)] \approx 0.015, \  
\mathbb{E}[u_A(x_1, y_1)] \approx 0.0.
\)
The  highest expected utility for player \( A \) occurs at the pair \( (x_1, y_2) \), where both agents play slightly biased distributions. Suppose now that neither agent can improve their outcome by unilaterally switching prompts due to their constrained prompt space. In this case, the pair \( (x_1, y_2) \) forms a \emph{reasoning equilibrium} in pure strategies, even though the induced behavior deviates from the classical Nash equilibrium. Specifically, the equilibrium behavioral strategies are:
\(
\mu_A^* = (0.2, 0.6, 0.2), \quad \mu_D^* = (0.3, 0.4, 0.3),
\)
which do not correspond to the uniform distribution \( (1/3, 1/3, 1/3) \). It  highlights that reasoning equilibria can exist outside the classical equilibrium set due to the constraints of the agents’ mindset.  

This game can be played repeatedly over time. Without expanding the agent's mindset, it is difficult to reach the classical Nash equilibrium \( (1/3, 1/3, 1/3) \). Instead, during the learning process, the agent must acquire new knowledge and enrich its reasoning space—i.e., expand its mindset—before the classical equilibrium can be approximated or achieved. This calls for an \emph{epistemic learning process}, wherein the agent not only updates its behavioral policy but also evolves its cognitive scaffolding \cite{cross2024hypothetical}. Such an epistemic process can be enabled by \emph{neurosymbolic learning} \cite{lei2023neurosymbolic}, which integrates neural learning mechanisms (e.g., from LLMs or reinforcement learning) with symbolic reasoning components. Neurosymbolic approaches allow agents to generalize beyond their initial prompt set, discover higher-order reasoning patterns, and formulate new prompts or abstractions that expand the space of attainable strategies. Over time, this can lead to more expressive mindsets and improved strategic outcomes that approach or even surpass classical equilibria in complex environments.

%Mindset will lead to a non-classical equilibrium.

\vspace{-2mm} \section{Conclusions} 

We have proposed an LLM-Nash game framework in which agents strategically select reasoning processes, represented by natural language prompts, to induce behavioral strategies that guide their actions. The equilibrium concept is thus defined not merely over the space of observable behaviors, but also over the underlying cognitive scaffolding that generates those behaviors. In this framework, the agents' reasoning processes are mediated by LLMs, and the space of accessible prompts defines their \emph{mindset}. The richness of a mindset governs both the level of rationality and the agent's \emph{strategic expressiveness}.

This shift from classical utility maximization to LLM-enabled, prompt-mediated reasoning closely mirrors human cognition, where decision-making proceeds from stimuli through perception, working memory, and mental reasoning before culminating in action. The LLM-Nash formulation is especially relevant in domains where bounded rationality, human behavior, and epistemic uncertainty are central. By formalizing reasoning explicitly, this framework allows for the analysis of how agents \emph{think}, not merely how they \emph{act}.

Future research directions include the development of a reasoning-level counterpart to correlated equilibrium. When agents share the same foundation model, parts of their cognitive structure and reasoning space are inherently aligned, introducing a natural form of correlation. This opens the door to new equilibrium concepts that unify LLM-mediated cognition and Bayesian game representations. Furthermore, the epistemic dimension and the associated learning processes offer a fertile avenue for exploration within this framework.

\vspace{-2mm}\bibliography{refs}

\begin{thebibliography}{10}
\providecommand{\url}[1]{\texttt{#1}}
\providecommand{\urlprefix}{URL }
\providecommand{\doi}[1]{https://doi.org/#1}

\bibitem{akerlof1978market}
Akerlof, G.A.: The market for “lemons”: Quality uncertainty and the market
  mechanism. In: Uncertainty in economics, pp. 235--251. Elsevier (1978)

\bibitem{bacsar2011prices}
Ba{\c{s}}ar, T., Zhu, Q.: Prices of anarchy, information, and cooperation in
  differential games. Dynamic Games and Applications  \textbf{1}(1),  50--73
  (2011)

\bibitem{blackwell1953equivalent}
Blackwell, D.: Equivalent comparisons of experiments. The Annals of
  Mathematical Statistics  \textbf{24}(2),  265--272 (1953).
  \doi{10.1214/aoms/1177729032}

\bibitem{cross2024hypothetical}
Cross, L., Xiang, V., Bhatia, A., Yamins, D.L., Haber, N.: Hypothetical minds:
  Scaffolding theory of mind for multi-agent tasks with large language models.
  arXiv preprint arXiv:2407.07086  (2024)

\bibitem{huang2023cognitive}
Huang, L., Zhu, Q.: Cognitive Security: A System-Scientific Approach. Springer
  (2023), forthcoming or update the year/publisher if officially released

\bibitem{lei2023neurosymbolic}
Lei, H., Zhu, Q.: Neurosymbolic meta-reinforcement lookahead learning achieves
  safe self-driving in non-stationary environments. arXiv preprint
  arXiv:2309.02328  (2023)

\bibitem{li2022role}
Li, T., Zhao, Y., Zhu, Q.: The role of information structures in game-theoretic
  multi-agent learning. Annual Reviews in Control  \textbf{53},  296--314
  (2022)

\bibitem{li2023price}
Li, T., Zhu, Q.: On the price of transparency: A comparison between overt
  persuasion and covert signaling. In: 2023 62nd IEEE Conference on Decision
  and Control (CDC). pp. 4267--4272. IEEE (2023)

\bibitem{liu2024exploring}
Liu, X., Zhang, J., Shang, H., Guo, S., Yang, C., Zhu, Q.: Exploring prosocial
  irrationality for llm agents: A social cognition view. arXiv preprint
  arXiv:2405.14744  (2024)

\bibitem{wickens2013engineering}
Wickens, C.D., Hollands, J.G., Banbury, S., Parasuraman, R.: Engineering
  Psychology and Human Performance. Pearson Education Inc., Upper Saddle River,
  NJ, USA, 4th edn. (2013)

\end{thebibliography}

\end{document}